\pdfoutput=1
\documentclass[11pt]{article}

\PassOptionsToPackage{numbers,sort&compress}{natbib}
\usepackage[preprint]{neurips_2025}
\usepackage{times}
\usepackage{latexsym}
\usepackage[T1]{fontenc}
\usepackage[utf8]{inputenc}
\usepackage[activate=false]{microtype}
\usepackage{inconsolata}
\usepackage{graphicx}
\usepackage{booktabs}
\usepackage{multirow}
\usepackage{amsmath, amssymb, amsfonts}
\usepackage{algorithm}
\usepackage{algorithmic}
\usepackage[font=small,labelfont=bf]{caption}
\usepackage{listings}
\usepackage{listingsutf8}
\usepackage[hidelinks]{hyperref}
\usepackage[nameinlink]{cleveref}
\usepackage{hyperref}

\usepackage{setspace}
\hypersetup{
  pdftitle={Aryabhata 2: Scaling Reinforcement Learning for Advanced STEM Reasoning},
  pdfauthor={Ritvik Rastogi, Tejas Chaudhari, Vishal Singh, Sandeep Varma}
}
\hypersetup{
    colorlinks=true,
    linkcolor=blue,
    filecolor=magenta,
    urlcolor=blue,
}

\newcommand\blfootnote[1]{%
  \begingroup
  \renewcommand\thefootnote{}\footnote{#1}%
  \addtocounter{footnote}{-1}%
  \endgroup
}

\title{Aryabhata 2: Scaling Reinforcement Learning for Advanced STEM Reasoning}

\author{Ritvik Rastogi \\
  PhysicsWallah \\
  \texttt{ritvik.rastogi@pw.live} \\\And
  Vishal Singh \\
  PhysicsWallah \\
  \texttt{vishal.singh16@pw.live} \\\And
  Tejas Chaudhari \\
  PhysicsWallah \\
  \texttt{tejas.chaudhari@pw.live} \\\And  
  Sandeep Varma \\
  PhysicsWallah \\
  \texttt{sandeep.varma@pw.live}
  }

\begin{document}
\maketitle

\begin{abstract}
\blfootnote{ \hspace{-0.65cm} Correspondence to \texttt{ritvik.rastogi@pw.live}.\\ Model available at \url{https://huggingface.co/PhysicsWallahAI/Aryabhata-2.0}.}

Competitive STEM examinations such as JEE and NEET require multi-step symbolic reasoning, precise numerical computation, and deep conceptual understanding across physics, chemistry, and mathematics. Recent large language models perform strongly on common reasoning benchmarks, yet they remain difficult to deploy at scale, where millions of student doubts demand domain-specific, consistently structured problem solving.

We introduce \textbf{Aryabhata 2}, a reasoning-focused language model for competitive STEM examinations, trained via reinforcement-learning post-training. Using PhysicsWallah's internal question banks, we construct a high-quality training curriculum and post-train GPT-OSS-20B through reinforcement learning with verifiable rewards. Training combines prolonged reinforcement learning with broadened exploration via progressively larger rollout group sizes. 

We evaluate Aryabhata 2 on competitive examination benchmarks, including JEE Main, JEE Advanced, and NEET, as well as out-of-distribution reasoning datasets such as AIME, HMMT, MMLU-Pro, MMLU-Redux 2.0, and GPQA. Results show that Aryabhata 2 outperforms its base model GPT-OSS-20B on competitive STEM reasoning while requiring substantially fewer output tokens (up to 64\% fewer).
\end{abstract}

\section{Introduction}
Competitive examinations such as the Joint Entrance Examination (JEE) and the National Eligibility cum Entrance Test (NEET)~\citep{jee_main_official,neet_ug_official} represent some of the most demanding reasoning tasks encountered in large-scale educational systems. Solving these problems requires multi-step symbolic manipulation, precise numerical reasoning, and deep conceptual understanding across physics, chemistry, and mathematics. Unlike many standard reasoning benchmarks, competitive exam questions are carefully constructed to test conceptual depth and often require chaining multiple reasoning steps under strict constraints.

PhysicsWallah’s online classes generate millions of student doubts, creating a large-scale requirement for fast, reliable, and clear STEM reasoning support. A substantial fraction of these doubts are tied to competitive examination preparation, where students expect not just final answers but step-by-step explanations that are accurate, concise, and aligned with exam-solving strategies.

This setting exposes a practical gap in current large language models (LLMs). Although recent models perform strongly on internet-scale and benchmark-style evaluations~\citep{deepseekr1_2025,qwen3technical2025}, they remain infeasible to deploy at scale due to high inference costs. For open-source models, these costs are driven by large model sizes and long chain-of-thought reasoning, while frontier models incur prohibitively high per-token pricing. This challenge is particularly acute for real student doubts, which require domain-specific syllabus coverage, multi-step symbolic reasoning, and strict correctness across physics, chemistry, and mathematics.

In this work, we introduce \textbf{Aryabhata 2}, a reasoning-focused language model trained specifically for competitive STEM examinations, building on our prior work Aryabhata~\citep{aryabhata1_2025}. Aryabhata 2 is obtained by post-training GPT-OSS-20B~\citep{gptoss2025}, an open-source 20B-parameter Mixture of Experts model with 3.6B activate parameters using reinforcement learning on a curated dataset derived from PhysicsWallah’s internal question banks covering Physics, Chemistry, Mathematics, and General Reasoning. Aryabhata 2 is openly released at \url{https://huggingface.co/PhysicsWallahAI/Aryabhata-2.0}~\citep{aryabhata2_hf}.

We evaluate Aryabhata 2 on competitive examination benchmarks including JEE Main, JEE Advanced, and NEET, as well as out-of-distribution reasoning datasets including AIME~\citep{aime_official}, HMMT~\citep{hmmt_official}, MMLU-Pro~\citep{mmlupro2024}, MMLU-Redux 2.0~\citep{mmluredux2025}, and GPQA~\citep{gpqa2023}. Our results demonstrate that targeted reinforcement learning on exam-style curricula can substantially improve reasoning performance on competitive STEM problems.

\section{Related Work}

Recent progress in reasoning-focused language models has increasingly relied on reinforcement learning (RL) during post-training. While supervised fine-tuning (SFT) remains the foundation for instruction-following behavior, reinforcement learning with verifiable rewards (RLVR) has proven particularly effective for domains where correctness can be programmatically evaluated, such as mathematics, coding, and symbolic reasoning. In such settings, reward signals derived from automatic verifiers enable scalable optimization beyond what is achievable with purely supervised datasets~\citep{deepseekmath2024,deepseekr1_2025}.

However, general reasoning spans heterogeneous domains with different supervision signals and verification costs. Mathematical problems often allow fast symbolic verification, while coding tasks require execution environments and subjective tasks rely on preference-based reward models. As a result, several distinct RL post-training paradigms have emerged.

\subsection{Sequential Reinforcement Learning}

Sequential RL pipelines apply reinforcement learning in staged curricula across domains. The \textit{Nemotron-Cascade} framework~\citep{nemotroncascade2025} proposes such a strategy, where models are optimized through a sequence of reinforcement learning stages including alignment, instruction following, mathematical reasoning, and coding tasks.

This approach treats post-training as a curriculum where general behaviors are learned before specialized reasoning capabilities. Sequential pipelines also offer practical engineering benefits since different domains can be trained with domain-specific infrastructure and verifier latency constraints. However, performance may depend on the ordering of stages, and poorly aligned objectives may lead to regressions in previously learned capabilities.

\subsection{Decentralized Training via Model Merging}

Another paradigm decomposes general reasoning into capability-specific experts that are trained independently and later combined through parameter merging. This approach has been explored in systems such as \textit{Command A}~\citep{commanda2025}, where multiple domain-specialized models are trained separately and merged via linear weight averaging.

Model merging enables parallel development of domain expertise and allows post-hoc rebalancing of capabilities without retraining. However, merged models may exhibit behavioral inconsistencies, often requiring additional alignment stages to produce a coherent policy.

\subsection{Unified Multi-Domain Reinforcement Learning}

Unified reinforcement learning approaches train models across multiple domains simultaneously within a single RL loop. The \textit{Nemotron 3 Nano}~\citep{nemotron3nano2025} training pipeline demonstrates this paradigm by exposing models to a mixture of reasoning environments including mathematics, coding, structured output tasks, and tool use.

Joint optimization across tasks helps mitigate catastrophic forgetting, since no domain is absent from training for extended periods. However, unified RL requires more complex infrastructure to handle heterogeneous reward functions and verification environments.

\subsection{Scaling Reinforcement Learning}

Recent work has also explored scaling reinforcement learning along new dimensions. \textbf{Prolonged Reinforcement Learning (ProRL)}~\citep{prorl2025} shows that reasoning performance can continue improving when RL training is extended to thousands of optimization steps, challenging the assumption that RL quickly reaches a performance plateau.

Complementary work on \textbf{Broadened Reinforcement Learning (BroRL)}~\citep{brorl2025} demonstrates that increasing the number of sampled rollouts per prompt can significantly improve exploration during training. By expanding the set of candidate reasoning trajectories, larger rollout groups increase the probability of discovering high-reward reasoning strategies.

\section{Methodology}

This section describes the full training pipeline used for Aryabhata 2, including data curation, answer verification, curriculum construction, and reinforcement learning. Our goal is to build a model for advanced STEM reasoning under practical compute constraints, by maintaining strong data quality and stable optimization.

Our approach uses a unified RL-only training pipeline. We begin with rigorous data cleaning and answer verification to ensure reliable reward supervision, then organize the verified corpus into a difficulty-aware curriculum. Training is performed in phased on-policy reinforcement learning: an initial format-alignment stage, a prolonged optimization stage for sustained capability gains, and a broadened exploration stage with larger rollout groups. This design combines the stability benefits of careful data curation with the performance gains of prolonged and broadened RL under constrained compute.

\subsection{Base Model}

Aryabhata 2 is built on top of \textbf{GPT-OSS-20B}, an open-source 20B-parameter model released by OpenAI~\citep{gptoss2025}. This model serves as the initial policy and is adapted using reinforcement learning.

To ensure training efficiency under limited hardware constraints, we use \textbf{parameter-efficient fine-tuning with Low-Rank Adaptation (LoRA)}~\citep{lora2021} instead of updating all model parameters. LoRA adapters are inserted into attention projection layers and the token embedding layer. This design substantially reduces memory usage while preserving enough adaptation capacity to improve reasoning behavior through reinforcement learning.

\subsection{Data Preparation}
\label{sec:data_preparation}

\subsubsection{Source Dataset}

The training dataset is constructed from PhysicsWallah’s internal question banks covering multiple STEM domains relevant to competitive examinations in India. These include questions from \textbf{Physics, Chemistry, Mathematics, and General Reasoning}, reflecting the distribution typically encountered in examinations such as JEE Main, JEE Advanced, and NEET.

The raw dataset initially contains \textbf{1.78M} questions. After applying a multi-stage cleaning pipeline and answer verification procedure, the dataset is reduced to \textbf{1.25M} high-quality questions that form the basis of our reinforcement learning curriculum.

To reduce benchmark contamination risk, we enforce a \textbf{knowledge cutoff of mid-2024} for the training corpus and apply decontamination against all held-out evaluation suites used in this work.

The distribution of questions across subjects at different stages of the pipeline is summarized in Table~\ref{tab:data_breakdown}.

\begin{table}[h]
\centering
\begin{tabular}{lcccc}
\hline
Subject & Raw & After Cleaning & After Verification & Final Curriculum \\
\hline
Physics & 471K & 342K & 314K & 30K \\
Chemistry & 527K & 375K & 345K & 30K \\
Mathematics & 534K & 419K & 385K & 30K \\
General Reasoning & 247K & 230K & 210K & 10K \\
\hline
\textbf{Total} & \textbf{1.78M} & \textbf{1.36M} & \textbf{1.25M} & \textbf{100K} \\
\hline
\end{tabular}
\caption{Dataset distribution across different stages of the preprocessing pipeline.}
\label{tab:data_breakdown}
\end{table}

\subsubsection{Question Cleaning Pipeline}

To ensure dataset reliability, we design a deterministic cleaning pipeline that removes malformed or unsuitable questions before the reinforcement learning stage.

First, \textbf{HTML artifacts are removed} from the dataset. In particular, questions containing \texttt{<img>} tags are discarded, since such questions typically depend on diagrams or figures that cannot be interpreted reliably by a text-only language model.

Second, we perform \textbf{LaTeX validation}. Many questions contain mathematical expressions written in LaTeX. To detect malformed expressions, we attempt to render all questions using \texttt{pdflatex}. Questions that fail compilation are discarded. This step ensures that the model receives syntactically valid mathematical content during training.

Third, we detect \textbf{incomplete or ill-posed questions} using a language model classifier. Specifically, we prompt \textit{Qwen/Qwen3-30B-A3B-Thinking-2507}~\citep{qwen3technical2025} to determine whether a question lacks sufficient information to be solved. Questions classified as incomplete are removed.

Finally, we apply \textbf{domain filtering} to remove non-STEM questions that do not belong to Physics, Chemistry, Mathematics, or General Reasoning categories using an instruction-tuned filtering model~\citep{qwen3technical2025}.

Across all stages, approximately \textbf{24\%} of the original dataset is removed.

\subsection{Answer Verification}

During preliminary inspection of the dataset, we observed that some of the original answer keys were incorrect or inconsistent. Since reinforcement learning relies on reward signals derived from answer correctness, such errors can significantly degrade training quality.

To address this issue, we design an automated answer verification pipeline based on two large language models.

The verification system consists of:

\begin{itemize}
\item \textbf{Policy model}: GPT-OSS-120B~\citep{gptoss2025}
\item \textbf{Judge model}: Qwen/Qwen3-30B-A3B-Thinking-2507~\citep{qwen3technical2025}
\end{itemize}

The policy model generates chain-of-thought (CoT) solutions~\citep{chainofthought2022} with temperature set to $1.0$, while the judge model evaluates whether the final answer of generated solution matches the ground truth.

\subsubsection{Multi-Pass Verification}

To maximize verification coverage while controlling computational cost, we employ a multi-stage sampling procedure.

\paragraph{Single-Sample Stage}

A single chain-of-thought solution is generated. If the judge model marks the answer as correct, the question-answer pair is accepted. Approximately \textbf{80\%} of the dataset is verified at this stage.

\paragraph{Four-Sample Stage}

For the remaining 20\% of questions, we generate four independent chain-of-thought solutions with different random seeds. A question is accepted if \textbf{any} of the generated solutions is judged correct. This stage verifies an additional \textbf{8\%} of the dataset.

\paragraph{Sixteen-Sample Stage}

For the remaining unresolved questions, we generate sixteen independent solutions and again accept the question if any solution is judged correct. This final stage verifies an additional \textbf{4\%} of the dataset.

The judge model provides binary correctness signals and questions are accepted if at least one valid reasoning trajectory exists.

\subsection{Curriculum Construction}

Once answer verification is complete, we construct a training curriculum based on empirical difficulty estimation.

For each question, we sample four independent generations at temperature $1.0$ using different random seeds. Based on the number of correct generations, questions are categorized into three difficulty levels:

\begin{itemize}
\item Trivial (4 out of 4 correct generations)
\item Learnable (1--3 out of 4 correct generations)
\item Challenging (0 out of 4 correct generations)
\end{itemize}

This categorization provides a natural measure of question difficulty relative to the base model's capabilities. We exclude trivial questions from the prolonged and broadened reasoning phases because they provide minimal learning signal. A small trivial subset is retained only for Phase 1 format alignment, and we retain challenging questions for the third phase of training.

During initial experiments, we observed that the base model performed significantly worse on \textbf{chemistry questions}. To mitigate this imbalance, chemistry questions are \textbf{upsampled} within the reinforcement learning curriculum.

\subsection{Unified Reinforcement Learning}

To improve reasoning ability, Aryabhata 2 is trained using an \textbf{on-policy reinforcement learning framework built upon Group Relative Policy Optimization (GRPO)}~\citep{deepseekmath2024}.

\subsubsection{Parameter-Efficient Adaptation}

Training uses Low-Rank Adaptation (LoRA), which enables parameter-efficient fine-tuning by introducing a small set of trainable parameters while keeping the base model weights frozen. This approach significantly reduces memory footprint and training cost while maintaining strong adaptation performance on domain-specific tasks.

\textbf{LoRA configuration:}
Table~\ref{tab:lora_config} summarizes the key hyperparameters.

\textbf{Parameter efficiency:}
Table~\ref{tab:param_efficiency} reports the total and trainable parameter counts.

In early ablation experiments, adding adapters to the token embedding layer significantly improved learning capacity.

\begin{center}
\begin{minipage}[t]{0.455\linewidth}
\vspace{0pt}
\centering
\begin{tabular}{l p{0.55\linewidth}}
    \toprule
    \textbf{Hyperparameter} & \textbf{Value} \\
    \midrule
    Rank ($r$) & 64 \\
    Scaling factor ($\alpha$) & 128 \\
    Dropout & 0 \\
    Target modules & $q\_proj$, $k\_proj$, $v\_proj$, $o\_proj$, $embed\_tokens$ \\
    \bottomrule
\end{tabular}
\captionof{table}{LoRA hyperparameters used for reinforcement learning post-training.}
\label{tab:lora_config}
\end{minipage}\hfill
\begin{minipage}[t]{0.455\linewidth}
\vspace{0pt}
\centering
\begin{tabular}{l c}
    \toprule
    \textbf{Metric} & \textbf{Value} \\
    \midrule
    Total parameters & 20,959,661,632 \\
    Trainable parameters & 31,850,496 \\
    Trainable parameters (\%) & 0.15\% \\
    \bottomrule
\end{tabular}
\vspace{7.5ex}
\captionof{table}{Total and trainable parameter counts for Aryabhata 2 with LoRA adapters.}
\label{tab:param_efficiency}
\end{minipage}
\end{center}

\subsubsection{RL Algorithm}

We use \textbf{GRPO as the base reinforcement learning algorithm}. For each prompt, we sample a group of responses and compute advantages relative to group reward statistics.

Compared to standard token-level GRPO implementations, we make the following modifications:

\begin{itemize}
\item \textbf{KL-free training:} We remove KL regularization and do not use a reference model. This design is motivated by GPU memory limits: keeping both policy and reference models exceeded the available memory on our two-H100 setup.

\item \textbf{DAPO-style clipped objective:} We optimize a clipped policy-ratio objective with an asymmetric upper clipping threshold~\citep{deepseekr1_2025}.

\item \textbf{No variance normalization:} Advantages are computed by subtracting the mean reward within each sampled group, without standard-deviation scaling.

\item \textbf{Truncation masking:} Completions that reach the maximum generation length are masked during optimization to avoid learning from incomplete trajectories.

\item \textbf{Multiplicative reward composition:} We use a product-form reward, $R = R_{accuracy} \times R_{format}$, to jointly enforce correctness and response-format quality.

\end{itemize}

\subsubsection{Reward Function}

As mentioned above, the final reward is computed multiplicatively:

\[
R = R_{accuracy} \times R_{format}
\]

\paragraph{Accuracy reward.}
The accuracy term is computed using an ordered matching cascade and depends on the question type.
The cascade proceeds through the following base matchers:

\begin{enumerate}
    \item \textbf{Case-insensitive string equality} after trimming whitespace.
    
    \item \textbf{Numeric matching with tolerance:} For numeric answers, let $a$ denote the parsed model prediction and $b$ denote the parsed gold value after whitespace stripping, \LaTeX{} wrapper removal, and scientific-notation normalization. We mark a numeric match when
    \[
    |a - b| \le \max\left(0.01 \cdot \max(|a|, |b|),\, 0.01\right).
    \]
    
    \item \textbf{Symbolic equivalence} via \textit{math-verify} after light \LaTeX{} normalization (with numeric fallback when symbolic verification fails).
\end{enumerate}

These matchers are applied sequentially, with the first successful match determining correctness.
Their usage varies slightly depending on the question format, as described below:

\begin{itemize}
    \item \textbf{True/false questions:} We apply raw and \LaTeX-stripped string matching.
    
    \item \textbf{Numerical, fill-in, and type-in questions:} We apply the full string--numeric--symbolic cascade.
    
    \item \textbf{Choice-style questions} (single-correct, multiple-correct, assertion--reasoning, and matching-list): Gold options are normalized to canonical labels and matched accordingly. For single-correct MCQs, if label matching fails but the predicted option value matches the gold option text, we assign partial credit of \(0.5\). Thus, \(R_{\text{accuracy}} \in \{0, 0.5, 1\}\).
    
    \item \textbf{Reward system design:} In initial experiments, the deterministic rule-based reward system defined above covers almost all cases in practice and is used as the sole reward-evaluation mechanism during RL training.
\end{itemize}

\paragraph{Format reward.}
In early experiments with just the accuracy reward, we observed that the model often terminated with a correct answer immediately after completing its reasoning, without providing a sufficiently detailed explanation for the student. Conversely, unconstrained reasoning could lead to excessively long outputs or reasoning loops. To balance these behaviors, we design a format reward that encourages sufficiently informative final answers while maintaining a controlled proportion between reasoning and solution length.

The format term is computed from output structure using character-level heuristics. Each output is split at the end-of-thinking delimiter into a reasoning segment and a final-answer segment. Let \(c_{tot}\) be the total number of output characters, \(c_{sol}\) be the number of characters in the final-answer segment, and

\[
\rho = \frac{c_{sol}}{c_{tot}}.
\]

We define

\[
R_{format} = S_{len}(c_{sol}) \times S_{ratio}(\rho),
\]

with

\[
S_{len}(c_{sol}) =
\begin{cases}
0, & c_{sol} < 100 \\
0.6, & 100 \le c_{sol} < 250 \\
0.8, & 250 \le c_{sol} < 500 \\
1.0, & c_{sol} \ge 500,
\end{cases}
\]

\[
S_{ratio}(\rho) =
\begin{cases}
\rho/0.3, & \rho < 0.3 \\
1.0, & 0.3 \le \rho \le 0.7 \\
(1-\rho)/0.3, & \rho > 0.7.
\end{cases}
\]

If parsing fails (e.g., missing delimiter), we set \(R_{format}=0\).

Intuitively, \(S_{len}\) rewards sufficiently detailed final answers by increasing the score with solution length, while \(S_{ratio}\) encourages a balanced allocation between reasoning and answer segments. Together, these terms discourage both overly brief responses and disproportionately long reasoning chains.

\subsection{Training Phases}

Training proceeds in three sequential phases.

\subsubsection{Phase 1: Format Alignment}

The first phase consists of \textbf{300 reinforcement learning steps} with a group size of \textbf{8}. Training is performed on a format-mixed dataset with chemistry questions upsampled.

The goal of this stage is to align the model with the desired answering format before scaling reasoning difficulty.

\subsubsection{Phase 2: Prolonged Reinforcement Learning}

The second phase runs for approximately \textbf{5,000 steps}. The group size is gradually increased from \textbf{8 to 16}.

During this phase, the dataset mixture is \textbf{adaptively adjusted} based on model evaluation results. Question difficulty is increased when the model sustains an accuracy reward greater than \textbf{0.7} for around \textbf{20 consecutive optimization steps}.

To stabilize training, we perform \textbf{EMA-based checkpoint merging} when reward improvements plateau. Multiple previous checkpoints are merged using exponential moving averaging to reduce instability.

\subsubsection{Phase 3: Broadened Reinforcement Learning}

The final phase focuses on exploration and generalization. This stage runs for approximately \textbf{700 steps} and increases the group size from \textbf{64 to 128}.

Larger group sizes enable broader exploration of reasoning trajectories, allowing the model to discover alternative solution strategies.

\paragraph{Training configuration across phases.}
We summarize the key hyperparameters across different training phases in Table~\ref{tab:training_phases}.

\begin{center}
\begin{tabular}{l c c c}
    \toprule
    \textbf{Parameter} & \textbf{Phase 1} & \textbf{Phase 2} & \textbf{Phase 3} \\
    \midrule
    Steps & 300 & 5000 & 700 \\
    Group size & 8 & 8 $\rightarrow$ 16 & 64$\rightarrow$128 \\
    Batch size & 128 & 128 $\rightarrow$ 256  & 512 $\rightarrow$ 1024 \\
    \# Data points & \textasciitilde5K & \textasciitilde80K & \textasciitilde15K \\
    Learning rate & 1e-6 & 1e-6 & 1e-6 \\
    Max tokens & 4K & 4K & 4K \\
    Difficulty level & Trivial & Learnable & Challenging \\
    \bottomrule
\end{tabular}
\captionof{table}{Training hyperparameters across the three reinforcement learning phases.}
\label{tab:training_phases}
\end{center}

\subsection{Training Infrastructure}

All experiments are conducted on \textbf{two NVIDIA H100 NVL GPUs}. Reinforcement learning is performed using on-policy sampling with generation temperature set to \textbf{1.0} during both training and evaluation.

Model evaluation is performed every \textbf{50 steps} using a held-out validation set. The final checkpoint is selected based on the highest validation accuracy. We use stochastic decoding and sample $k=4$ responses per question, and compute Pass@1 as the mean correctness across sampled responses. Majority voting is not used in reported metrics.

\section{Evaluation}

We evaluate Aryabhata 2 on a suite of competitive examination datasets and established reasoning benchmarks. The evaluation is designed to measure performance both on \textbf{in-distribution exam-style problems} and \textbf{out-of-distribution reasoning benchmarks}.

\subsection{Metrics}
\label{sec:eval_metrics}

We default to stochastic pass@$k$ evaluation (rather than greedy decoding) and report Pass@1 using non-zero-temperature sampling~\citep{codex2021}. For each question, we sample $k$ responses (with $k=4$ in our main evaluation), and compute
\[
\mathrm{Pass@1} = \frac{1}{k} \sum_{i=1}^{k} p_i,
\]
where $p_i \in \{0,1\}$ denotes the correctness of the $i$-th response.

In addition to raw accuracy, we report the \textbf{accuracy--token trade-off}, which measures accuracy relative to the number of output tokens generated during inference. This metric captures practical deployment efficiency for real-world, large-scale tutoring and question-answering systems. We compute \textbf{accuracy per 1K output tokens} as
\[
\text{Acc./1K tokens} = \frac{\text{Pass@1 (4-sample mean)}}{\text{output tokens}} \times 1000,
\]
where Pass@1 and output tokens are averaged across benchmarks within each split.

\subsection{Benchmarks}

\subsubsection{In-Distribution Benchmarks}

Evaluation is conducted on recent competitive examinations. These exams require a combination of conceptual understanding, multi-step reasoning, and precise numerical computation across physics, chemistry, and mathematics.
All benchmarks are restricted to text-only questions, excluding those that require diagrams or visual interpretation. The benchmark composition is summarized in Table~\ref{tab:in_distribution_benchmarks}.
NEET includes Biology questions; since Biology is not part of the RL curriculum described in Section~\ref{sec:data_preparation}, we treat this as a partial distribution shift within the in-distribution exam suite and report aggregate results for operational comparability.

\begin{center}
\begin{tabular}{l l c}
    \toprule
    Dataset & Subject & \# Questions \\
    \midrule
    \multirow{4}{*}{\textbf{JEE Main 2025}} 
        & Physics & 435 \\
        & Chemistry & 344 \\
        & Mathematics & 475 \\
        & \textbf{Total} & \textbf{1254} \\
    \midrule
    \multirow{4}{*}{\textbf{JEE Advanced 2025}} 
        & Physics & 12 \\
        & Chemistry & 22 \\
        & Mathematics & 29 \\
        & \textbf{Total} & \textbf{63} \\
    \midrule
    \multirow{4}{*}{\textbf{NEET 2025}} 
        & Physics & 33 \\
        & Chemistry & 33 \\
        & Biology & 84 \\
        & \textbf{Total} & \textbf{150} \\
    \midrule
    \multirow{4}{*}{\textbf{JEE Main 2026 (January Session)}} 
        & Physics & 169 \\
        & Chemistry & 171 \\
        & Mathematics & 217 \\
        & \textbf{Total} & \textbf{557} \\
    \bottomrule
\end{tabular}
\captionof{table}{In-distribution benchmark datasets and question counts used for evaluation.}
\label{tab:in_distribution_benchmarks}
\end{center}

\subsubsection{Out-of-Distribution Benchmarks}

To measure generalization beyond the training distribution, we evaluate on a diverse set of widely used reasoning benchmarks spanning olympiad-style mathematics and broad STEM knowledge: AIME~\citep{aime_official}, HMMT~\citep{hmmt_official}, MMLU-Pro~\citep{mmlupro2024}, MMLU-Redux 2.0~\citep{mmluredux2025}, and GPQA~\citep{gpqa2023}. The dataset breakdown is provided in Table~\ref{tab:ood_benchmarks}.

\begin{center}
\begin{tabular}{l c}
    \toprule
    Dataset & \# Questions \\
    \midrule
    AIME (2025--2026) & 60 \\
    HMMT (2025--2026) & 93 \\
    MMLU-Pro (PCMB) & 4,500 \\
    MMLU-Redux 2.0 (High School and College PCMB) & 763 \\
    GPQA & 546 \\
    \bottomrule
\end{tabular}
\captionof{table}{Out-of-distribution benchmark datasets and question counts used to evaluate generalization.}
\label{tab:ood_benchmarks}
\end{center}

The evaluation spans both competition-level mathematics (AIME, HMMT) and broad STEM reasoning (MMLU-Pro~\citep{mmlupro2024}, MMLU-Redux 2.0~\citep{mmluredux2025}, and GPQA~\citep{gpqa2023}), providing a comprehensive measure of generalization.

\subsection{Baselines}

We compare Aryabhata 2 against a mixture of open-source reasoning models and frontier proprietary models.

\paragraph{Open-source models.}
We include recent open-weight models with strong reasoning capabilities: Qwen3-30B-A3B (Thinking)~\citep{qwen3technical2025}, Nemotron 3 Nano 30B A3B~\citep{nemotron3nano2025}, GPT-OSS-20B, and GPT-OSS-120B~\citep{gptoss2025}.

\paragraph{Frontier models.}
We additionally evaluate against frontier proprietary models included in our result tables: GPT-5 Mini and GPT-5 Nano~\citep{gpt5_2025}, and Gemini 2.5 Flash~\citep{gemini25flash_2025}.

\subsection{Answer Verification}

To determine answer correctness, we apply a multi-stage answer extraction and matching pipeline:

\begin{enumerate}
\item \textbf{String Matching:} case-insensitive exact matching after whitespace normalization.
\item \textbf{Numeric Matching:} tolerance-based numerical equivalence checks after whitespace normalization, \LaTeX{} wrapper stripping, and scientific-notation normalization.
\item \textbf{Symbolic Matching:} symbolic equivalence using the \textit{math-verify} library after light \LaTeX{} normalization.
\item \textbf{Option Matching:} canonical-label matching for choice-style questions.
\item \textbf{LLM-as-Judge:} if previous steps fail, an LLM extracts the final answer from the response and compares it to the ground truth while ignoring intermediate reasoning steps.
\end{enumerate}

This pipeline ensures robust evaluation across numeric, symbolic, and multiple-choice answers while minimizing extraction errors from generated reasoning traces.

\subsection{Results}

\subsubsection{In-Distribution Results}

Table~\ref{tab:in_distribution_results} reports overall Pass@1 (4-sample mean) on the in-distribution exam benchmarks. Aryabhata 2 achieves the strongest open-source aggregate performance, with an average of \textbf{88.95}, compared with 88.28 for GPT-OSS-120B, 88.55 for Qwen3-30B-A3B (Thinking), and 83.00 for GPT-OSS-20B.

In addition to accuracy gains, Aryabhata 2 is substantially more token-efficient than GPT-OSS-20B on in-distribution exams, reducing output tokens by approximately 52--64\% across the four datasets.

Aryabhata 2 achieves \textbf{42.31} Acc./1K tokens. This is substantially higher than GPT-OSS-20B (15.68), GPT-OSS-120B (26.66), Nemotron 3 Nano 30B A3B (14.41), and Qwen3-30B-A3B (Thinking) (19.44). Detailed in-distribution accuracy--token values for all models are listed in Appendix Table~\ref{tab:acc_token_id_appendix}.

\begin{center}
\small
\resizebox{\linewidth}{!}{
\begin{tabular}{lccccc}
    \toprule
    Model & JEE Adv. 2025 & NEET 2025 & JEE Main 2025 & JEE Main 2026 & Avg. \\
    \midrule
    GPT-5 Mini & 93.65 & 87.33 & 87.07 & 95.83 & 89.71 \\
    GPT-5 Nano & 84.13 & 82.33 & 78.58 & 81.68 & 79.89 \\
    Gemini 2.5 Flash & 96.81 & 90.00 & 87.26 & 96.22 & 90.23 \\
    \midrule
    Qwen3-30B-A3B (Thinking) & 90.48 & 86.00 & 84.89 & 97.26 & 88.55 \\
    Nemotron 3 Nano 30B A3B & 90.87 & 84.00 & 82.89 & 94.84 & 86.51 \\
    GPT-OSS-120B & 84.13 & 85.33 & 85.61 & 95.42 & 88.28 \\
    GPT-OSS-20B & 77.38 & 81.33 & 79.27 & 92.46 & 83.00 \\
    \midrule
    Aryabhata 2 (ours) & 86.51 & 84.66 & 87.80 & 92.99 & 88.95 \\
    \bottomrule
\end{tabular}
}
\captionof{table}{In-distribution Pass@1 (4-sample mean, \%) overall accuracy. Avg. is the average across the four in-distribution benchmarks.}
\label{tab:in_distribution_results}
\end{center}

\subsubsection{Out-of-Distribution Results}

Table~\ref{tab:ood_results} summarizes OOD performance on Olympiad-style mathematics and broad STEM reasoning benchmarks using Pass@1 (4-sample mean). Aryabhata 2 attains an OOD average of \textbf{87.64}, improving over GPT-OSS-20B (84.95) and Nemotron 3 Nano 30B A3B (83.48), while remaining below GPT-5 Mini (88.85), Gemini 2.5 Flash (89.13), Qwen3-30B-A3B (Thinking) (89.42), and GPT-OSS-120B (89.50).

Compared with GPT-OSS-20B, Aryabhata 2 matches AIME performance (86.67), improves on HMMT (+1.54), GPQA (+4.35), and MMLU-Pro (+3.07), and shows a small decline on MMLU-Redux 2.0 (-0.40). Against Qwen3-30B-A3B-Thinking, Aryabhata 2 shows a large gain on HMMT (+27.08), indicating stronger robustness on this harder Olympiad-style benchmark.

Aryabhata 2 is also more token-efficient than GPT-OSS-20B across all OOD benchmarks, reducing output tokens by approximately 24--71\%.

Aryabhata 2 attains \textbf{39.58} Acc./1K tokens. This improves over GPT-OSS-20B (17.48), GPT-OSS-120B (24.44), Nemotron 3 Nano 30B A3B (13.61), and Qwen3-30B-A3B (Thinking) (20.80). The full OOD accuracy--token summary is reported in Appendix Table~\ref{tab:acc_token_ood_appendix}.

Figures~\ref{fig:acc_token_in_distribution} and~\ref{fig:acc_token_ood} show accuracy versus output tokens for in-distribution and out-of-distribution averages, respectively.

\begin{center}
\small
\resizebox{\linewidth}{!}{
\begin{tabular}{lcccccc}
    \toprule
    Model & AIME & HMMT & GPQA & MMLU-Pro & MMLU-Redux 2.0 & Avg. \\
    \midrule
    GPT-5 Mini & 83.33 & 70.97 & 75.46 & 89.64 & 96.40 & 88.85 \\
    GPT-5 Nano & 74.17 & 63.98 & 61.11 & 80.62 & 88.47 & 79.51 \\
    Gemini 2.5 Flash & 66.61 & 59.13 & 75.09 & 90.44 & 96.85 & 89.13 \\
    \midrule
    Qwen3-30B-A3B (Thinking) & 84.58 & 51.88 & 73.31 & 90.80 & 97.77 & 89.42 \\
    Nemotron 3 Nano 30B A3B & 77.08 & 65.86 & 65.38 & 84.33 & 94.10 & 83.48 \\
    GPT-OSS-120B & 90.00 & 80.01 & 77.06 & 90.11 & 95.94 & 89.50 \\
    GPT-OSS-20B & 86.67 & 77.42 & 70.51 & 85.42 & 93.32 & 84.95 \\
    \midrule
    Aryabhata 2 (ours) & 86.67 & 78.96 & 74.86 & 88.49 & 92.92 & 87.64 \\
    \bottomrule
\end{tabular}
}
\captionof{table}{Out-of-distribution Pass@1 (4-sample mean, \%) accuracy. Avg. is the average across five OOD benchmarks.}
\label{tab:ood_results}
\end{center}

\begin{center}
\includegraphics[draft=false,width=0.95\linewidth]{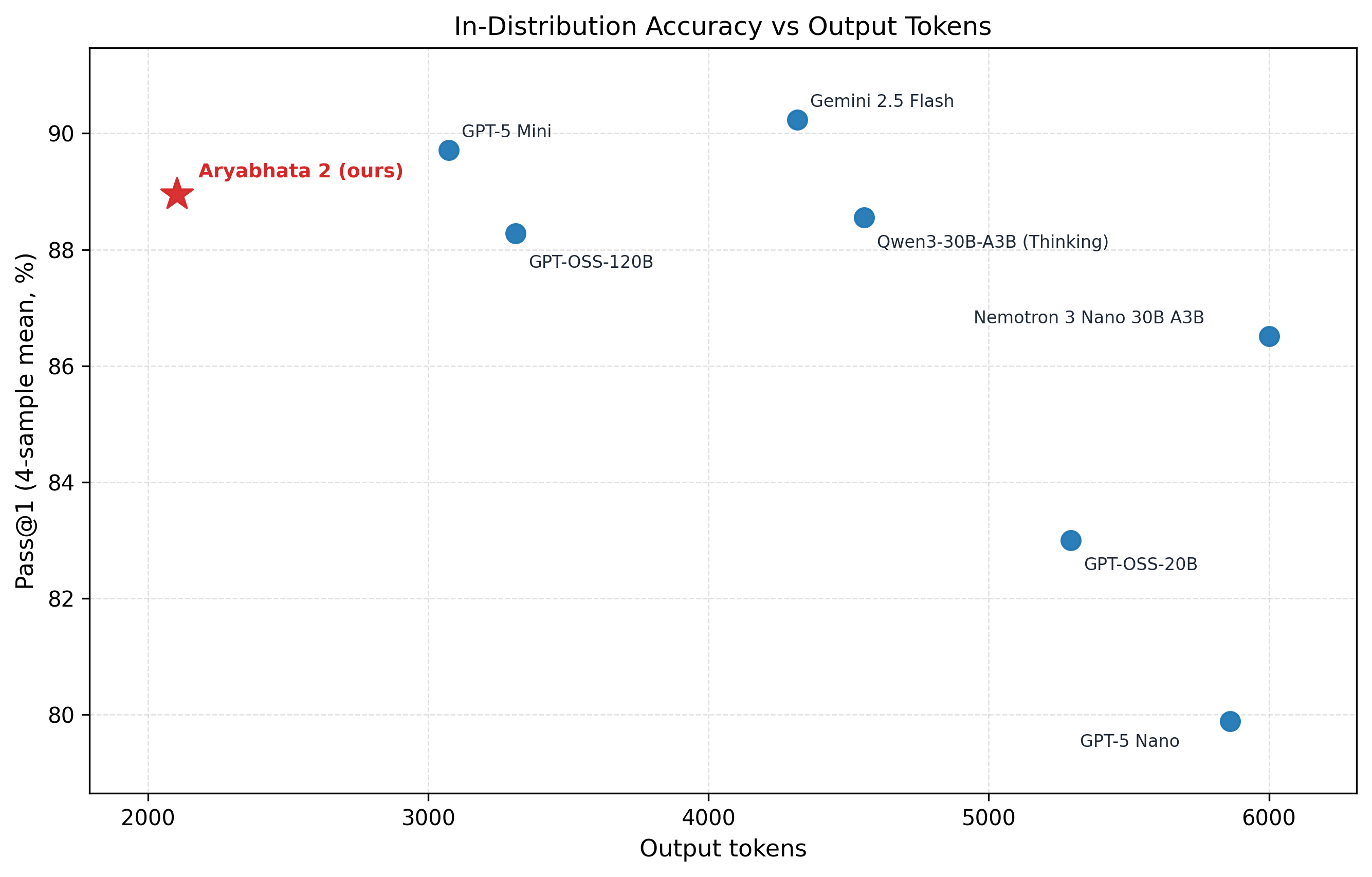}
\captionof{figure}{In-distribution accuracy-token trade-off (y-axis: Pass@1 (4-sample mean), x-axis: output tokens). Each point is a model-level average across the in-distribution benchmarks.}
\label{fig:acc_token_in_distribution}
\end{center}

\begin{center}
\includegraphics[draft=false,width=0.95\linewidth]{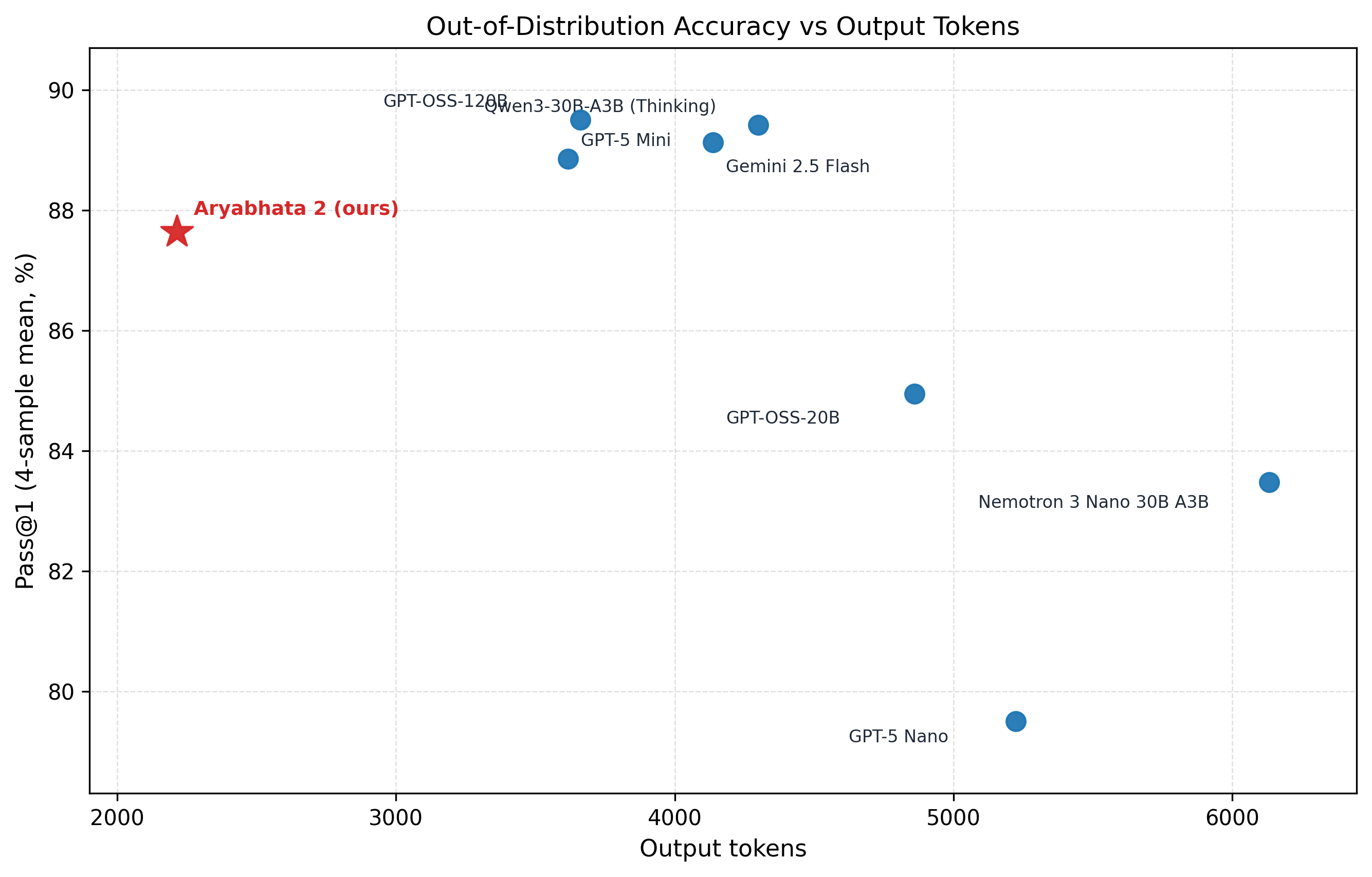}
\captionof{figure}{Out-of-distribution accuracy-token trade-off (y-axis: Pass@1 (4-sample mean), x-axis: output tokens). Each point is a model-level average across the OOD benchmarks.}
\label{fig:acc_token_ood}
\end{center}

\section*{Conclusion}

In this work, we presented \textbf{Aryabhata 2}, a reinforcement-learning post-trained 20B model designed for advanced competitive STEM reasoning. Our pipeline combines rigorous data cleaning and answer verification with phased reinforcement learning that includes format alignment, prolonged optimization, and broadened exploration. This design enables stable training under constrained compute while preserving strong reasoning quality on exam-style problems.

Across in-distribution benchmarks, Aryabhata 2 achieves strong performance, including a \textbf{92.99} score on \textbf{JEE Main 2026}. Relative to the base GPT-OSS-20B model, Aryabhata 2 improves overall accuracy on all in-distribution exams and delivers substantially shorter generations. On out-of-distribution benchmarks, Aryabhata 2 improves over GPT-OSS-20B in Pass@1 (4-sample mean) while remaining competitive with larger baselines. The accuracy-token analysis further shows favorable deployment efficiency, particularly on in-distribution tasks, where Aryabhata 2 attains \textbf{88.95} Pass@1 at \textbf{2,102.25} output tokens and \textbf{42.31} Acc./1K tokens.

These results suggest that targeted RL on domain-specific curricula is an effective strategy for scaling practical STEM reasoning systems for real educational workloads. We release Aryabhata 2 on HuggingFace \href{https://huggingface.co/PhysicsWallahAI/Aryabhata-2.0}{(PhysicsWallahAI/Aryabhata-2.0)} to support further research on exam-focused language models.

\bibliographystyle{plainnat_keepcaps}
\bibliography{latex/main}
\onecolumn
\appendix
\raggedbottom

\section{Additional Evaluation Tables}

\subsection{Accuracy--Token Trade-off Tables}

\begin{center}
\small
\resizebox{\linewidth}{!}{
\begin{tabular}{lccc}
    \toprule
    Model & Pass@1 (\%) & Output tokens & Acc./1K tokens \\
    \midrule
    GPT-5 Mini & 89.71 & 3,072.79 & 29.20 \\
    GPT-5 Nano & 79.89 & 5,861.08 & 13.63 \\
    Gemini 2.5 Flash & 90.23 & 4,316.30 & 20.90 \\
    \midrule
    Qwen3-30B-A3B (Thinking) & 88.55 & 4,555.93 & 19.44 \\
    Nemotron 3 Nano 30B A3B & 86.51 & 6,001.59 & 14.41 \\
    GPT-OSS-120B & 88.28 & 3,311.56 & 26.66 \\
    GPT-OSS-20B & 83.00 & 5,293.04 & 15.68 \\
    \midrule
    Aryabhata 2 (ours) & 88.95 & 2,102.25 & 42.31 \\
    \bottomrule
\end{tabular}
}
\captionof{table}{In-distribution accuracy--token trade-off summary.Pass@1 and output tokens are averaged across JEE Advanced 2025, NEET 2025, JEE Main 2025, and JEE Main 2026.}
\label{tab:acc_token_id_appendix}
\end{center}

\begin{center}
\small
\resizebox{\linewidth}{!}{
\begin{tabular}{lccc}
    \toprule
    Model & Pass@1 (\%) & Output tokens & Acc./1K tokens \\
    \midrule
    GPT-5 Mini & 88.85 & 3,616.77 & 24.57 \\
    GPT-5 Nano & 79.51 & 5,222.71 & 15.22 \\
    Gemini 2.5 Flash & 89.13 & 4,137.27 & 21.54 \\
    \midrule
    Qwen3-30B-A3B (Thinking) & 89.42 & 4,299.45 & 20.80 \\
    Nemotron 3 Nano 30B A3B & 83.48 & 6,132.90 & 13.61 \\
    GPT-OSS-120B & 89.50 & 3,661.45 & 24.44 \\
    GPT-OSS-20B & 84.95 & 4,859.79 & 17.48 \\
    \midrule
    Aryabhata 2 (ours) & 87.64 & 2,214.35 & 39.58 \\
    \bottomrule
\end{tabular}
}
\captionof{table}{Out-of-distribution accuracy--token trade-off summary.Pass@1 and Output tokens are averaged across AIME, HMMT, GPQA, MMLU-Pro, and MMLU-Redux 2.0.}
\label{tab:acc_token_ood_appendix}
\end{center}

\subsection{Subject-wise In-Distribution Accuracy}

\begin{center}
\small
\resizebox{\linewidth}{!}{
\begin{tabular}{lcccc}
    \toprule
    Model & Math & Physics & Chemistry & Overall \\
    \midrule
    GPT-5 Mini & 99.14 & 83.33 & 92.05 & 93.65 \\
    GPT-5 Nano & 98.28 & 75.00 & 70.45 & 84.13 \\
    Gemini 2.5 Flash & 100.00 & 91.60 & 95.45 & 96.81 \\
    \midrule
    Qwen3-30B-A3B (Thinking) & 87.93 & 87.50 & 95.45 & 90.48 \\
    Nemotron 3 Nano 30B A3B & 93.10 & 87.50 & 89.77 & 90.87 \\
    GPT-OSS-120B & 96.55 & 58.33 & 81.82 & 84.13 \\
    GPT-OSS-20B & 92.24 & 50.00 & 72.73 & 77.38 \\
    \midrule
    Aryabhata 2 (ours) & 93.10 & 70.83 & 86.36 & 86.51 \\
    \bottomrule
\end{tabular}
}
\captionof{table}{JEE Advanced 2025 subject-wise Pass@1 (4-sample mean, \%).}
\label{tab:subjectwise_jee_adv_2025_appendix}
\end{center}

\begin{center}
\small
\resizebox{\linewidth}{!}{
\begin{tabular}{lcccc}
    \toprule
    Model & Biology & Physics & Chemistry & Overall \\
    \midrule
    GPT-5 Mini & 80.95 & 100.00 & 90.91 & 87.33 \\
    GPT-5 Nano & 77.98 & 95.45 & 80.30 & 82.33 \\
    Gemini 2.5 Flash & 84.52 & 100.00 & 93.93 & 90.00 \\
    \midrule
    Qwen3-30B-A3B (Thinking) & 80.95 & 100.00 & 84.85 & 86.00 \\
    Nemotron 3 Nano 30B A3B & 77.38 & 96.97 & 87.88 & 84.00 \\
    GPT-OSS-120B & 82.14 & 90.91 & 87.88 & 85.33 \\
    GPT-OSS-20B & 75.00 & 96.97 & 81.82 & 81.33 \\
    \midrule
    Aryabhata 2 (ours) & 80.95 & 96.96 & 81.82 & 84.66 \\
    \bottomrule
\end{tabular}
}
\captionof{table}{NEET 2025 subject-wise Pass@1 (4-sample mean, \%).}
\label{tab:subjectwise_neet_2025_appendix}
\end{center}

\begin{center}
\small
\resizebox{\linewidth}{!}{
\begin{tabular}{lcccc}
    \toprule
    Model & Math & Physics & Chemistry & Overall \\
    \midrule
    GPT-5 Mini & 94.97 & 83.05 & 81.28 & 87.07 \\
    GPT-5 Nano & 93.89 & 75.17 & 61.77 & 78.58 \\
    Gemini 2.5 Flash & 96.42 & 80.16 & 83.59 & 87.26 \\
    \midrule
    Qwen3-30B-A3B (Thinking) & 93.53 & 79.00 & 80.41 & 84.89 \\
    Nemotron 3 Nano 30B A3B & 91.23 & 76.35 & 79.66 & 82.89 \\
    GPT-OSS-120B & 95.59 & 79.68 & 79.56 & 85.61 \\
    GPT-OSS-20B & 92.54 & 73.27 & 68.54 & 79.27 \\
    \midrule
    Aryabhata 2 (ours) & 95.21 & 84.25 & 82.04 & 87.80 \\
    \bottomrule
\end{tabular}
}
\captionof{table}{JEE Main 2025 subject-wise Pass@1 (4-sample mean, \%).}
\label{tab:subjectwise_jee_main_2025_appendix}
\end{center}

\begin{center}
\small
\resizebox{\linewidth}{!}{
\begin{tabular}{lcccc}
    \toprule
    Model & Math & Physics & Chemistry & Overall \\
    \midrule
    GPT-5 Mini & 97.93 & 95.27 & 93.71 & 95.83 \\
    GPT-5 Nano & 88.02 & 78.97 & 76.31 & 81.68 \\
    Gemini 2.5 Flash & 95.85 & 97.60 & 95.32 & 96.26 \\
    \midrule
    Qwen3-30B-A3B (Thinking) & 98.39 & 96.60 & 96.49 & 97.26 \\
    Nemotron 3 Nano 30B A3B & 97.93 & 93.34 & 92.40 & 94.84 \\
    GPT-OSS-120B & 97.35 & 93.49 & 94.88 & 95.42 \\
    GPT-OSS-20B & 96.54 & 92.31 & 87.43 & 92.46 \\
    \midrule
    Aryabhata 2 (ours) & 96.42 & 93.93 & 87.71 & 92.99 \\
    \bottomrule
\end{tabular}
}
\captionof{table}{JEE Main 2026 subject-wise Pass@1 (4-sample mean, \%).}
\label{tab:subjectwise_jee_main_2026_appendix}
\end{center}

\end{document}